\pdfoutput=1

\documentclass[11pt]{article}

\usepackage[]{naacl2021}

\usepackage{times}
\usepackage{booktabs}
\usepackage{amsmath}
\usepackage{amssymb}
\usepackage{latexsym}
\usepackage{verbatim}
\usepackage[T1]{fontenc}

\usepackage[utf8]{inputenc}

\usepackage{microtype}
\usepackage{graphicx}
\usepackage[normalem]{ulem}

\usepackage{tcolorbox}

\title{Learning to Ask Like a Physician}

\author{Eric Lehman\textsuperscript{1,2,\thanks{\texttt{ lehmer16@mit.edu}}},
Vladislav Lialin\textsuperscript{3}, 
Katelyn Y. Legaspi\textsuperscript{5},
Anne Janelle R. Sy\textsuperscript{4},\\
\textbf{Patricia Therese S. Pile\textsuperscript{4},
Nicole Rose I. Alberto\textsuperscript{5},
Richard Raymund R. Ragasa\textsuperscript{5}},\\ 
\textbf{Corinna Victoria M. Puyat\textsuperscript{5},
Isabelle Rose I. Alberto\textsuperscript{5},
Pia Gabrielle I. Alfonso\textsuperscript{5}},\\
\textbf{Marianne Taliño\textsuperscript{6},
Dana Moukheiber\textsuperscript{1},
Byron C. Wallace\textsuperscript{7},
Anna Rumshisky\textsuperscript{3}},\\
\textbf{Jennifer J. Liang\textsuperscript{2,8},
Preethi Raghavan\textsuperscript{1,9}, 
Leo Anthony Celi\textsuperscript{1, 10},
Peter Szolovits\textsuperscript{1,2}}\\
\textsuperscript{1}MIT, 
\textsuperscript{2}MIT-IBM Watson AI Lab, 
\textsuperscript{3}University of Massachusetts Lowell,
\textsuperscript{4}UERM Memorial\\ Medical Center
\textsuperscript{5}University of the Philippines,
\textsuperscript{6}ASMPH,
\textsuperscript{7}Northeastern University,\\
\textsuperscript{8}IBM Research,
\textsuperscript{9}Fidelity Investments,
\textsuperscript{10}Beth Israel Deaconess Medical Center
}

\begin{document}
\maketitle

\begin{abstract}
Existing question answering (QA) datasets derived from electronic health records (EHR) are artificially generated and consequently fail to capture realistic physician information needs.
We present \textbf{Di}scharge \textbf{S}ummary \textbf{C}linical \textbf{Q}uestions (DiSCQ), a newly curated question dataset composed of 2,000+ questions paired with the snippets of text (\emph{triggers}) that prompted each question. 
The questions are generated by medical experts from 100+ MIMIC-III discharge summaries. 
We analyze this dataset to characterize the types of information sought by medical experts. 
We also train baseline models for trigger detection and question generation (QG), paired with unsupervised answer retrieval over EHRs. Our baseline model is able to generate high quality questions in over 62\% of cases when prompted with human selected triggers.
We release this dataset (and all code to reproduce baseline model results) to facilitate further research into realistic clinical QA and QG. \footnote{\url{https://github.com/elehman16/discq}}
\end{abstract}

\section{Introduction}

Physicians often query electronic health records (EHR) to make fully informed decisions about patient care \cite{fushman2009}. 
However, \citet{Alessandro2004} found that 
it takes an average of 8.3 minutes to answer a single question, 
even when physicians are trained to retrieve information from an EHR platform. 
Natural language technologies such as automatic question answering (QA) may partially address this problem. 

There have been several dataset collection efforts that aim to facilitate the training and evaluation of clinical QA models \cite{pampari2018emrqa, yue2021cliniqg4qa, raghavan-etal-2021-emrkbqa, kell-etal-2021-take}. 
However, template-based \cite{pampari2018emrqa,raghavan-etal-2021-emrkbqa} and other kinds of automated generation \cite{yue2021cliniqg4qa} methods are by nature brittle and have limited evidence of producing questions that medical professionals ask. 

Datasets such as emrQA \cite{pampari2018emrqa} and emrKBQA \cite{raghavan-etal-2021-emrkbqa} attempt to simulate physician queries by defining templates derived from actual questions posed by physicians and then performing slot-filling with clinical entities.
This method yields questions that are structurally realistic, but not consistently medically relevant. 
\citet{yue-etal-2020-clinical} found that 
sampling just 5\% of the emrQA questions was sufficient for training a model. They further note that 96\% of the questions in a subsection of emrQA contain key phrases that overlap with those in the selected answer.

\begin{figure}
    \centering
    \includegraphics[width=0.5\textwidth]{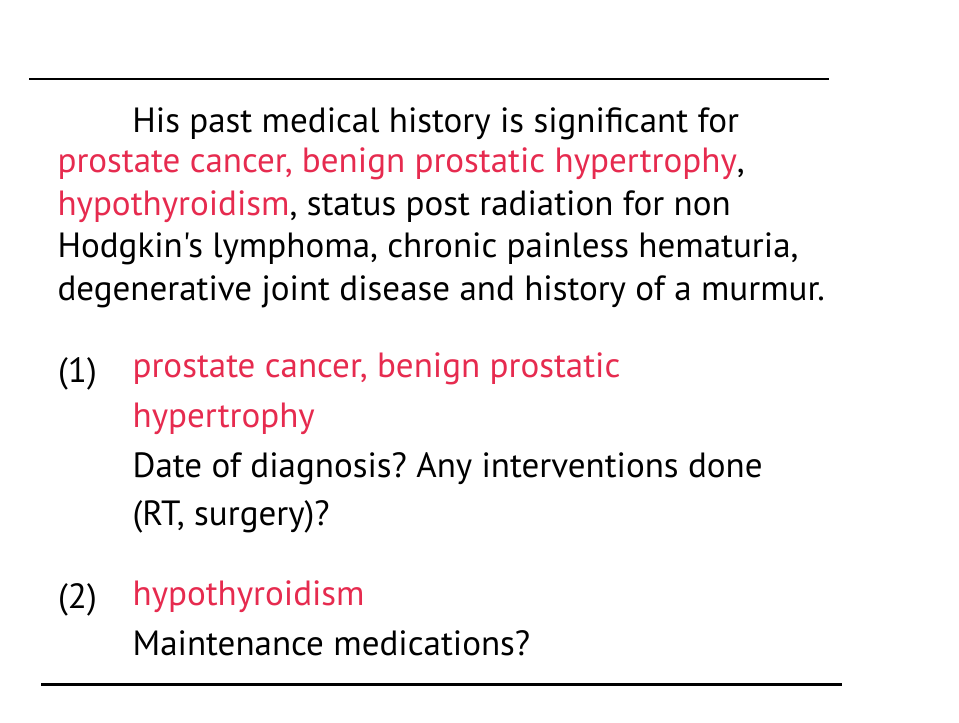}
    \caption{Example of an annotated discharge summary section. The highlighted portion shows the ``trigger'' for the questions.}
    \label{fig:ex_dataset}
\end{figure}


In follow-up work, \citet{yue2021cliniqg4qa} provide a new dataset of 975 questions generated using a diverse question generation model with a human-in-the-loop and 312 questions generated by medical experts from scratch, with the caveat that they must be answerable on the given discharge summary. However, a random sample of 100 questions from the former reveals that 96\% of the 975 questions were slot-filled templates directly from emrQA. A separate random sample of 100 questions from the latter set reveals that 54\% of the questions also use the same slot-filled templates from emrQA. 
Similarly, we find that 85\% of the machine-generated questions and 75\% of the human-generated questions contain the exact same key phrases as in the selected answer. 
Although \citet{yue-etal-2020-clinical} does not discuss how they prompt physician questions, our analysis strongly suggests that even in the case of questions ``written'' by physicians, answer spans are likely identified in advance; this significantly constrains the set of questions a medical professional can ask. 

To 
address this paucity of natural, clinically relevant questions, we collect queries that might plausibly be asked by healthcare providers during patient handoff (i.e., transitions of care). 
We use patient discharge summaries from the Medical Information Mart for Intensive Care III (MIMIC-III) English dataset \cite{mimiciii} to mimic the handoff process. 
We expect this process to produce more natural questions than prior work.
We work with 10 medical experts of varying skill levels.
We ask them to review a given discharge summary as the receiving physician in a patient handoff and record any questions they have as well as the piece of text within the discharge summary (trigger) that prompted the question.
A sample of questions and corresponding triggers can be seen in Figure \ref{fig:ex_dataset}.


We train question trigger detection and question generation (QG) models on DiSCQ, paired with unsupervised answer retrieval over the EHR.
Finally, we propose a new set of guidelines for human evaluation of clinical questions and evaluate the performance of our pipeline using these guidelines.
Concretely, our contributions 
are summarized as follows:


\begin{itemize}
    \item We work with 10 medical experts to compile DiSCQ, a new dataset of 2000+ questions and 1000+ triggers from over 100+ discharge summaries, providing an important new resource for research in clinical NLP.
    \item We demonstrate the dataset's utility by training baseline models for trigger detection and question generation.
    \item We develop novel guidelines for human evaluation of clinical questions.
    Our experiments show
    that widely used automated QG metrics do not correlate with human-evaluated question quality.
\end{itemize}

\section{Related Work}
\subsection{Clinical Question Datasets}
\label{sec:related-work:datasets}

\begin{figure*}
    \centering
    \noindent\makebox[\textwidth]{\includegraphics[width=1\textwidth]{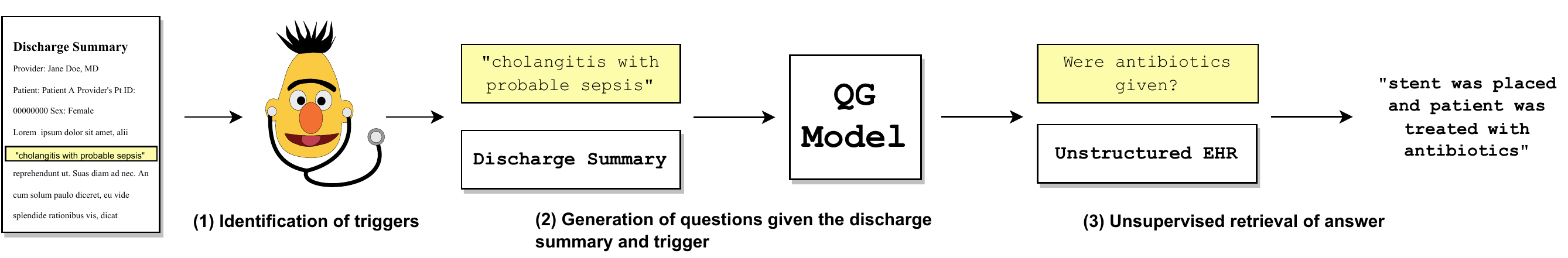}}
    \caption{Schematic of the pipeline process used to generate and answer questions.}
    \label{fig:qg_full_pipeline}
\end{figure*}

Clinical information retrieval, and in particular clinical question answering, is a challenging research task with direct potential applications in clinical practice. Several dataset collection efforts gather consumer health questions and pair them with answers from sources like WebMD and PubMED
%
\cite{Yu2007DevelopmentIA, Cao2011AskHERMESAO, Abacha2015MEANSAM, Abacha2017OverviewOT, Zahid2018CLINIQAAM, DemnerFushman2020ConsumerHI-CHQA, Savery2020QuestiondrivenSO-MEDIQA, zhu-etal-2020-mashqa, Abacha2019BridgingTG}.
Likewise, \citet{Suster2018CliCRAD-CLICR} automatically generate 100,000+ information retrieval queries from over 11,000+ BMJ Case Reports.
While these resources are helpful in testing a model's understanding and information retrieval ability on biomedical texts, these datasets consist of broad medical questions asked by the general population. Doctors will not only ask more specific and targeted questions, but also query the EHR to make fully informed decisions about patient care.



The number of publicly available QA datasets derived from EHR systems is 
quite limited due to the labor intensiveness and high skill requirement needed to create such a dataset. 
%
As mentioned previously, to help alleviate this dearth of clinical questions, \citet{pampari2018emrqa} introduced emrQA, a QA dataset constructed from templatized physician queries slot-filled with n2c2 annotations.\footnote{https://www.i2b2.org/NLP/DataSets/} 
\citet{fan-2019-annotating} extended emrQA by explicitly focusing on ``why'' questions.
\citet{Soni2019UsingFT} introduced a novel approach for constructing clinical questions that can be slot-filled into logical-forms.
\citet{yue2021cliniqg4qa} applied an emrQA-trained question generation model paired with a human-in-the-loop to collect 1287 questions conditioned on and answerable from the given context.

In contrast, 
in our data collection process 
we do not restrict the medical expert to ask only questions 
answerable from a particular part of the discharge summary. 
This leads to more diverse and natural questions. 
Additionally, in DiSCQ each question is associated with a span of text that triggered the question.

\subsection{Question Generation}
Question Generation (QG) is a challenging task that requires a combination of reading comprehension and text generation. 
Successful QG models may aid in education \cite{heilman-smith-2010-good, DuSC17}, creating dialogue systems or chatbots \cite{ShangLL15, mostafazadeh-etal-2016-corpus, shum2018eliza}, building datasets \cite{duan-etal-2017-question} or improving question answering models through data augmentation \cite{TangDQZ17, dong2019unified, puri-etal-2020-training, yue2021cliniqg4qa}. 

Most QG approaches can be broken down into either rule-based or neural methods. Rule-based approaches often involve slot filling templatized questions \cite{heilman-smith-2010-good, mazidi-nielsen-2014-linguistic, labutov-etal-2015-deep, chali-hasan-2015-towards, pampari2018emrqa}. While often effective at generating numerous questions, these methods are very rigid, as virtually any domain change requires a new set of rules.
This problem is particularly important in medical QG, as different types of practices may focus on varying aspects of a patient and therefore ask different questions.

Compared to rule-based methods, sequence-to-sequence models \cite{SerbanKTTZBC16, DuSC17} and more recently transformer-based models \cite{dong2019unified, qi-etal-2020-prophetnet, lelkes2021quizstyle, Murakhovska2021MixQGNQ, Luo2021CooperativeLO} allow for generation of more diverse questions and can potentially mitigate the problem of domain generalization via large-scale pre-training \cite{brown2020gpt3} or domain adaptation techniques. We choose to train both BART \cite{lewis-etal-2020-bart} and T0 \cite{sanh2021multitask} models for the task of question generation due to their high performance and ability to generalize to new tasks.

\section{DiSCQ Dataset}
\label{dataset_section}
We work with 10 medical experts of varying skill levels, ranging from senior medical students to practicing MDs, to 
construct a dataset of 2029 questions over 100+ discharge summaries from MIMIC-III \cite{mimiciii}.

\subsection{Dataset Collection}
The goal of our question collection is to gather questions that may be asked by healthcare providers during patient handoff (i.e., transitions of care). We use the patient discharge summary to simulate the handoff process,\footnote{We discard any records pertaining to neonatal or deceased patients.} where the discharge summary is the communication from the previous physician regarding the patient’s care, treatment and current status. Annotators are asked to review the discharge summary as the receiving physician and ask any questions they may have as the physician taking over the care of this patient.

Annotators are instructed to read the discharge summary line-by-line and record (1) any questions that may be important with respect to the patient's future care, and, (2) the text within the note that triggered the question. This may mean that questions asked early on may be answered later in the discharge summary. Annotators are permitted to go back and ask questions if they feel the need to do so. To capture the annotators' natural thought processes, we purposely provide only minimal guidance to annotators on how to select a trigger or what type of questions to ask. We only ask that annotators use the minimum span of text when specifying a trigger.\footnote{Instructions given to annotators will be available \href{https://github.com/elehman16/discq}{here}.}

We also encourage all questions to be asked in whatever format they feel appropriate. This leads to many informal 
queries, in which 
questions are incomplete or grammatically incorrect (Figure~\ref{fig:ex_dataset}). 
Further, we encourage all types of questions to be asked, regardless of whether they could be answered based on the EHR. 
We also allow the annotators to ask an arbitrary number of questions.
This allows for annotators to skip discharge summaries entirely should they not have any questions.

\subsection{Dataset Statistics}


The trigger/question pairs are generated over entire discharge summaries. We instruct annotators to select the minimum span that they used as the trigger to their question; this leads to triggers of length $5.0 \pm 14.1$ tokens. We additionally find that there are $1.86 \pm 1.56$ questions per trigger.
As mentioned previously, we encourage our medical experts to ask questions however they feel 
most comfortable. 
This led to a wide variety in how questions were asked, with some entirely self-contained (46\%), others requiring the trigger for understanding (46\%), and some requiring the entire sentence containing the trigger to 
comprehend (8\%).\footnote{Based on a sample of 100 questions.} 
We also observe that 59\% of the bi-grams in our questions are unique (i.e., over half of all bi-grams that appear in one question are not seen in any other question), demonstrating the diversity of how our questions are asked (Table \ref{tab:dataset-comp}).

\begin{figure}
    \centering
    \hspace*{-2mm} 
    \includegraphics[width=0.50\textwidth]{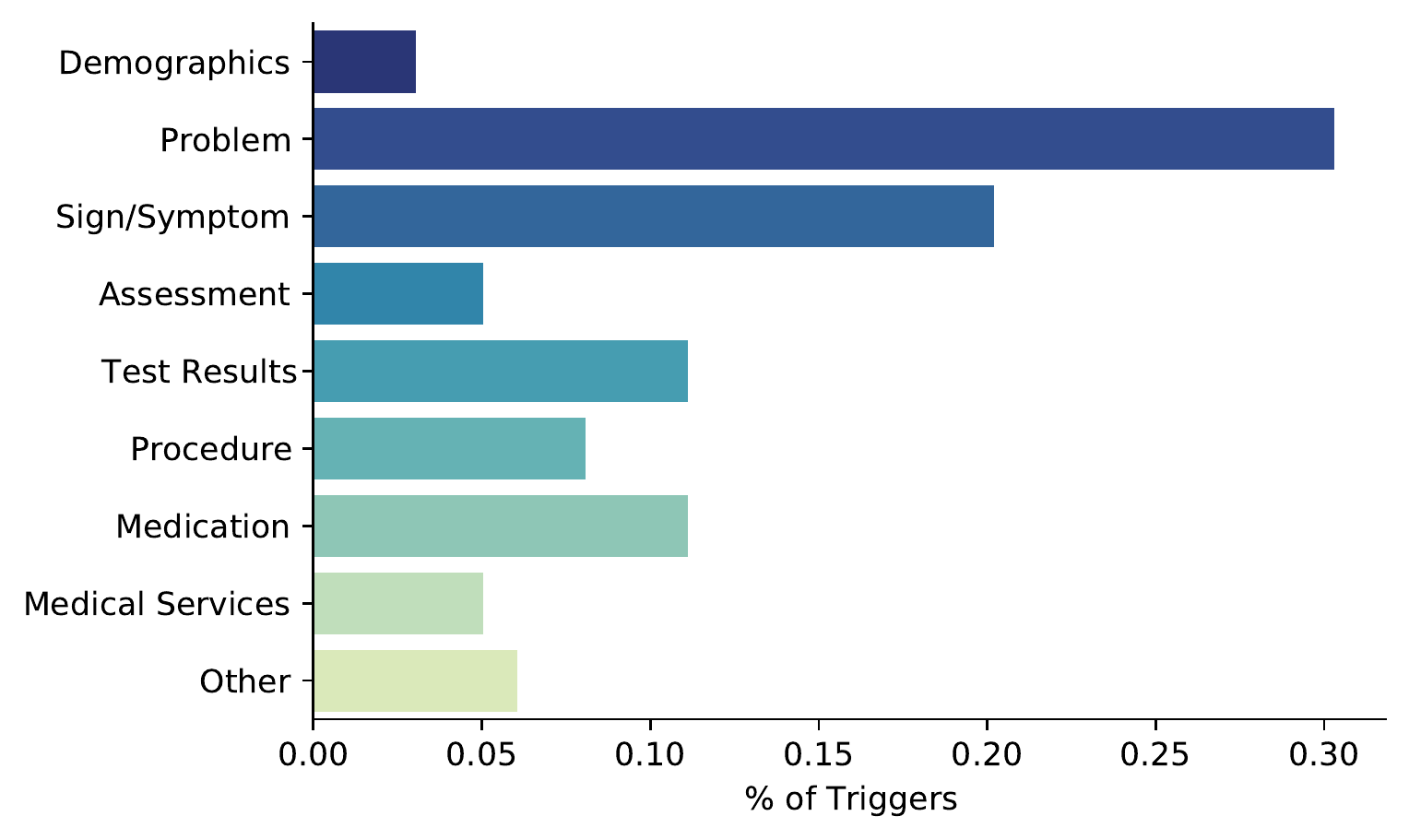}
    \caption{We randomly sample 100 gold triggers and have one of the authors, a physician, categorize the type of information that the trigger contains.}
    \label{fig:distribution_of_trigger_type}
\end{figure}

We additionally examine where in the discharge summary annotators 
tend to select triggers from. We find that a majority of triggers are selected from the \texttt{Hospital Course} (13\%) and \texttt{History of Present Illness} (39\%) sections. This is unsurprising, as these are the narrative sections of the note where the patient's history prior to admission and their medical care during hospitalization are described. 
Further, we find that a majority of triggers selected are either a \texttt{Problem} or \texttt{Sign/Symptom} (Figure \ref{fig:distribution_of_trigger_type}). 
This aligns with our intuition, as clinicians are often trained to organize patient information from a problem-oriented perspective. Moreover, developing a differential diagnosis usually begins with gathering details of the patient's clinical presentation.

\begin{figure}
    \centering
    \hspace*{-0.5cm} 
    \includegraphics[width=0.5\textwidth]{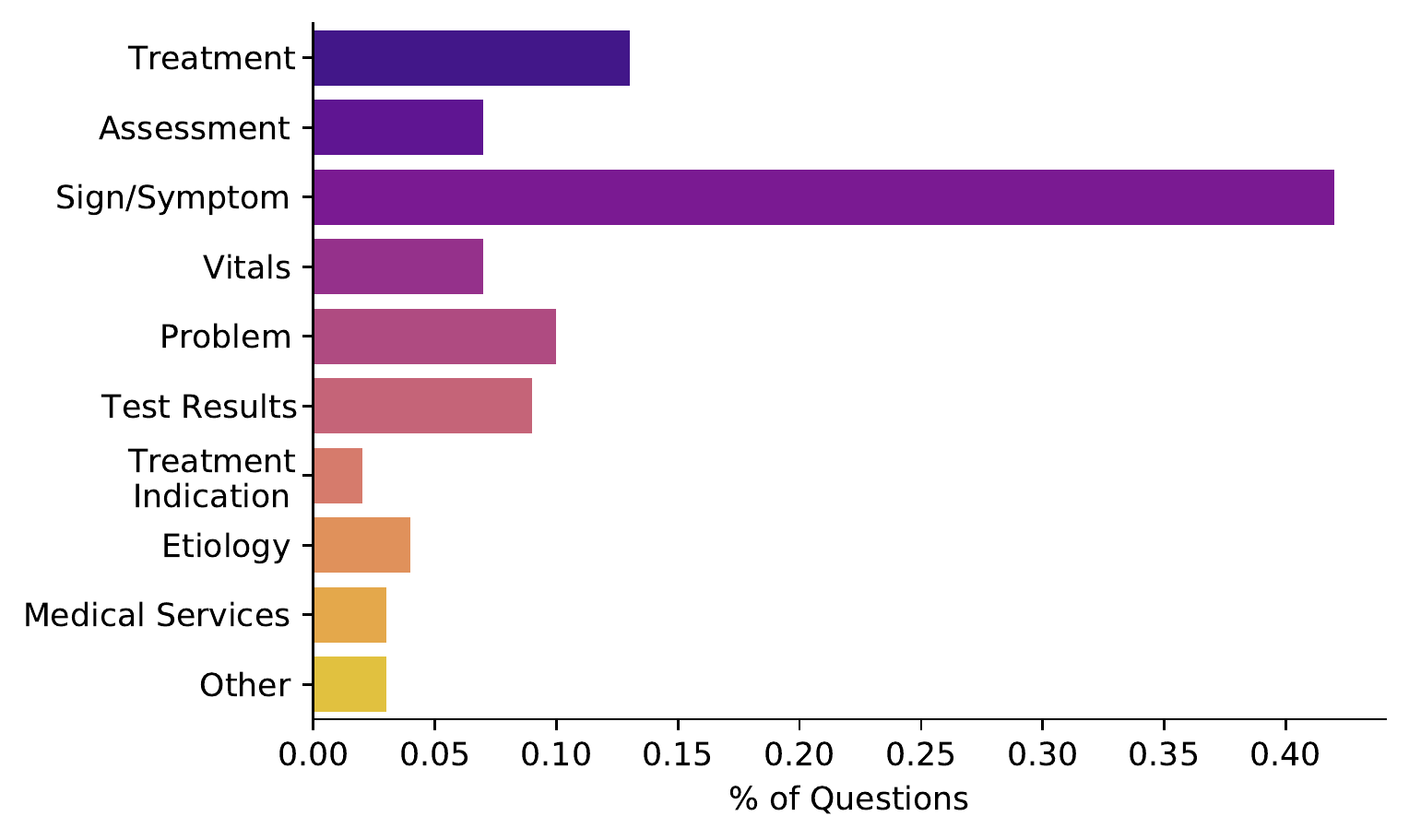}
    \caption{We randomly sample 100 questions and have one of the authors, a physician, categorize what type of information the question is asking for.}
    \label{fig:distribution_of_info_requests}
\end{figure}

In Figure \ref{fig:distribution_of_info_requests}, we examine the types of information needs exhibited by our questions. 
\begin{table}[h]
    \centering
    \small
    \begin{tabular}{p{1in} lll}
    \hline
        \textbf{Characteristics} &  \textbf{emrQA} &  \textbf{CliniQG4QA} &  \textbf{DiSCQ} \\\hline
        Total Articles  & 2,425 & 36 & 114 \\
        Total Questions & 455,837 & 1287 & 2029 \\
        Questions / Article & 187 & 35.8 & 17.8 \\
        Article Length & 3828 & 2644 & 1481 \\
        Question Length & 7.8 & 8.7 & 4.4 \\
        Unique Question \\Bi-grams & - & 24\% & 59\% \\
        Physician Generated & 0\% & 24\% & 100\% \\ 
        Indicates Question \\Motivation & No & No & Yes \\
        \hline
    \end{tabular}
    \caption{Comparison of emrQA, CliniQG4QA and our dataset. Question and article length scale given in tokens. Unique question bi-grams is given as a ratio.}
    \label{tab:dataset-comp}
\end{table}
We find that 83\% and 80\% of the questions categorized as \texttt{Sign/Symptom} and \texttt{Problem}, respectively, stem from the same category of trigger. \texttt{Sign/Symptom} questions generated from \texttt{Sign/Symptom} triggers are usually asking about associated symptoms (e.g., Trigger: \textit{dysuria}; Question: \textit{Any perineal rash or irritation?}) 
or additional details about the trigger (e.g., onset, timing).
Similarly, \texttt{Problem} questions generated from \texttt{Problem} triggers are usually asking about associated comorbid conditions or additional details of a diagnosis (e.g., date of diagnosis, severity).
We interestingly find that 62\% of the \texttt{Treatment} questions and 56\% of the \texttt{Test Results} questions are derived from triggers of type \texttt{Problem}. 
This can be attributed to diagnostic tests being used to monitor disease progression and treatment questions asking about how a problem is managed. 

As a soundness check, we randomly sample 100 questions from our dataset and find that only 22\% of them directly map to emrQA templates. 
Of the 22 that match, 17 of them map directly to \texttt{|problem|?} and \texttt{|test|?}. 
Additionally, we sample 100 questions to determine where a physician would hypothetically search the EHR should they choose to find the answers to these questions.\footnote{We use the same sample of 100 questions as before.} 
We find that one of the authors, a physician, would search external resources 3\% of the time, the structured data 20\% of the time and both the structured and unstructured data 21\% of the time. The remaining 56\% of questions would be answered solely from unstructured EHR data. This differs significantly from both emrQA and CliniQG4QA, in which all questions are answerable using unstructured EHR data.


As mentioned previously, we provide only minimal guidance on how to select a trigger or what type of question to ask, in order to capture the annotators' natural thought processes. 
The task is purposely presented in an open-ended fashion to encourage natural questions.
This may lead to situations in which two annotators examining the same discharge summary focus on entirely different aspects of the patient.
Such a scenario is likely to be common, as if most experts agree that a piece of information is important, then it would likely already be in the discharge summary.
We can attempt to measure this variation between medical experts by calculating trigger level agreement in documents annotated by two different annotators (roughly 50\% of discharge summaries in DiSCQ). We find a Cohen Kappa of 0.08.\footnote{This is calculated on a per-token level.} 

This lower agreement can be expected, as 
different spans can express the same information due to information redundancy in clinical notes.
Furthermore, clinical reasoning is not a linear process; therefore, different triggers can lead to the same question. For example, an expression of elevated blood pressure (\textit{"blood pressure of 148 to 162/45 to 54"}) and a diagnosis of hypertension (\textit{"Hypertension"}) led two annotators to both ask about the patient's normal blood pressure range.
We do not measure agreement of questions asked, as this is an inherently subjective task and questions are asked \textit{because} of differences between medical experts. 



\section{Task Setup}
We consider the task of generating questions that are relevant to a patient's care, given a discharge summary and a trigger. 
Afterwards, we attempt to find answers to these generated questions (Figure~\ref{fig:qg_full_pipeline}).
We also examine model performance for when the trigger is not provided and must instead be predicted. 
The task of generating questions without triggers can be viewed similarly to answer-agnostic question generation. 
We take a similar approach to \cite{subramanian-etal-2018-neural}, in which we implement a pipeline system that first selects key phrases from the passage and then generates questions about the selected key phrases.

While a majority of past works attempt to ensure that the generated question is answerable \cite{Nema2019LetsAA, pan-etal-2020-semantic, Wang_Wang_Tao_Zhang_Xu_2020, Huang_Fu_Mo_Cai_Xu_Li_Li_Leung_2021}, we do not impose this constraint.
In fact, we argue that the ability to generate unanswerable questions is necessary for real-world applications, as a question answering system
should be able to identify such questions. These questions can be used as hard-negatives to train and calibrate QA systems.

\section{Models}
\label{sec:models}

Pre-trained transformers have become ubiquitous in many natural language processing tasks \cite{devlin-etal-2019-bert,raffel2020T5,sanh2021multitask}, including natural language generation \cite{lewis-etal-2020-bart,Bao2020UniLMv2PL}.
Additionally, large-scale transformers have demonstrated the importance of parameter count for both upstream \cite{Kaplan2020ScalingLF} and downstream tasks, especially in low-resource settings \cite{brown2020gpt3,sanh2021multitask}.
As these results were mainly shown in non-clinical general domains, we find it important to evaluate both medium-sized and large models.

We formulate trigger detection as a tagging problem, for which we fine-tune ClinicalBERT \cite{alsentzer-etal-2019-publicly}.
For question generation, we experiment with both BART (406M parameters) \cite{lewis-etal-2020-bart} and T0 (11B parameters) \cite{sanh2021multitask}. Question generation is formulated as a conditional generation problem and modelled via a sequence-to-sequence approach. During evaluation, we use greedy sampling to produce generated text.

\paragraph{Reducing context size}
Due to memory constraints and the limited sequence length of pretrained models, we only select the part of the discharge summary containing the trigger. This is done in two possible ways: (1) extracting the sentence\footnote{Sentence splitting is performed using ScispaCy's \texttt{en\_core\_sci\_md}.} with the trigger or multiple sentences if a trigger spans across sentence boundaries or (2) extracting a chunk of size 512 containing the trigger in it.
To check if this context is actually used by the models we also fine-tune BART without extra discharge summary context (trigger text only).

\paragraph{Handling multiple questions}
41\% of the DiSCQ examples have multiple questions per trigger. Sometimes the questions depend on each other: 
\begin{itemize}
\itemsep0em 
    \item \textit{What meds was used? dosage? and route of administration?}
    \item \textit{Any culture done? What were the findings?}
\end{itemize}
For this reason, we train and evaluate models in two different setups: split questions (by the \texttt{?}-symbol) and combined questions. While the split-questions format might be more comparable to pre-existing work, the combined-questions setting likely models more realistic behavior of medical professionals.

\paragraph{Prompting}
\citet{Schick2021ItsNJ} demonstrate that adding natural language instructions to the model input can significantly improve model quality. The area of prompting has recently gained widespread popularity \cite{liu2021pretrain} and has had particular success in low-supervision scenarios \cite{Schick2021ItsNJ}. 
T0 \cite{sanh2021multitask} is a fine-tuned T5 \cite{raffel2020T5} model trained on 64 datasets and prompts from the Public Pool of Prompts \cite{bach2022promptsource}.
Given a trigger and some context from the discharge summary, we fine-tune T0++ and BART with the following prompt:
``\texttt{\{context\}}After reading the above EMR, what question do you have about "\texttt{\{trigger\}}"? Question:''.



\section{Results}
We split 2029 questions into train (70\%), validation (10\%) and test (20\%) sets\footnote{We use a document level split.} and fine-tune the models as described in Section \ref{sec:models}.
To evaluate trigger detection, we use token-level precision, recall and F1 score.
For automated evaluation of question generation we use ROUGE-L \cite{lin-2004-rouge}, METEOR \cite{banerjee-lavie-2005-meteor} and BERTScore \cite{zhang2020bertscore} metrics.
To monitor the diversity of generated questions, we measure the fraction of unique questions on the evaluation set.
%
As the question generation task has high variability of plausible generations, the utility of automatic metrics is debatable due to poor correlation with human evaluation \cite{callison-burch-etal-2006-evaluating, novikova-etal-2017-need, elliott-keller-2014-comparing, zhang2020bertscore, Bhandari2020ReevaluatingEI}. For this reason, we additionally perform human evaluation (Section \ref{sec:human-eval}).

\subsection{Trigger detection}
\label{sec:trigger_dection}
As mentioned in Section \ref{dataset_section}, we collect triggers for each question asked. We train a simple ClinicalBERT model to predict whether or not each token-piece is a trigger. To ground these results, we additionally use ScispaCy Large \cite{Neumann2019ScispaCyFA} to tag and classify all clinical entities as triggers. 
Results are shown in Table \ref{tab:trigger-id}.

\begin{table}[h]
    \centering
    \small
    \begin{tabular}{l lll}
    \hline
        Model            & Recall & Precision & F1    \\
        \hline
        ScispaCy         & 0.186  & 0.033     & 0.056 \\
        ClinicalBERT     & 0.184  & 0.196     & 0.190 \\
         \hline
    \end{tabular}
    \caption{Trigger detection results on the test set.
    }
    \label{tab:trigger-id}
\end{table}

We see that our model exhibits poor performance likely due to the fact that there is low agreement between annotators about which spans to highlight when asking questions. 

\subsection{Question generation}

Automated metrics for question generation experiments are available in Table \ref{tab:question_generation_automatic}.
While generation diversity changes significantly between different models, ranging from 30\% of unique questions to 79\%, METEOR, ROUGE-L and BERTScore show very similar and low performance across the board.


However, upon observation, many of the generated questions seem reasonable (Table \ref{tab:generations}), suggesting that these metrics might not fit the task. We hypothesize that this is caused by two reasons: (1) the short length of our questions and (2) a high number of potentially reasonable questions that could be generated. As we observe during the data collection process, different annotators seem to ask different questions despite citing the same trigger.
For these reasons, human evaluation (Section \ref{sec:human-eval}) might be a more appropriate approach for testing the quality of these models. 


\subsection{Answer Selection}
\label{sec:answer_selection}
In addition to identifying triggers and generating questions, we attempt to find answers to these questions. We only consider the unstructured portion of the EHR data. We train a ClinicalBERT model on emrQA augmented with unanswerable questions via negative sampling \cite{risk_score_work}. Due to the question's frequent dependency on the trigger, given a trigger and a question, we prompt the model with the following text: ``With respect to \texttt{\{trigger\}}, \texttt{\{question\}}?''. We first query the remainder of the discharge summary that the question was generated from. 
If we are unable to find an answer with probability above some threshold\footnote{This threshold was chosen manually by examining question-answer pairs on a validation set.}, we query the model on prior patient notes. 
We then select the highest probability span and expand it to a sentence level prediction. 
We always return a prediction even in cases where all sentences are equally unlikely to be the answer.
\begin{table*}[h]
\begin{small}
    \centering
    \begin{tabular}{p{63mm}|p{42mm}|p{17mm}|p{19mm}}
        \toprule
        Context & Generated Question & Trigger Type & Question Type  \\
        \midrule
        Pt reports that he noticed a \textit{right neck mass} last October & Size, outline (asymmetry), color, elevation, evolving? & sign/symptom & sign/symptom  \\
        \midrule
        She was also significantly \textit{tachypneic} & were there interventions done to address this? & sign/symptom & treatment \\
        \midrule
        According to Dr. <name>, she has had stable deficits for many years \textit{without any flare-like episodes}. & How is her vision now? & assessment & sign/symptom \\
        \midrule
        Her bicarb began to drop and she developed an \textit{anion gap acidosis} & confusion? confusion? agitation? hand tremors? bounding pulses? & problem & sign/symptom  \\
    \end{tabular}
    \caption{Example T0 model generations, cherry-picked. This model examines single sentences and is trained with combined questions. Trigger phrases are \textit{italicized}.}
    \label{tab:generations}
\end{small}
\end{table*}

\section{Human Evaluation}
\label{sec:human-eval}

\begin{figure}
    \centering
    \includegraphics[width=0.5\textwidth]{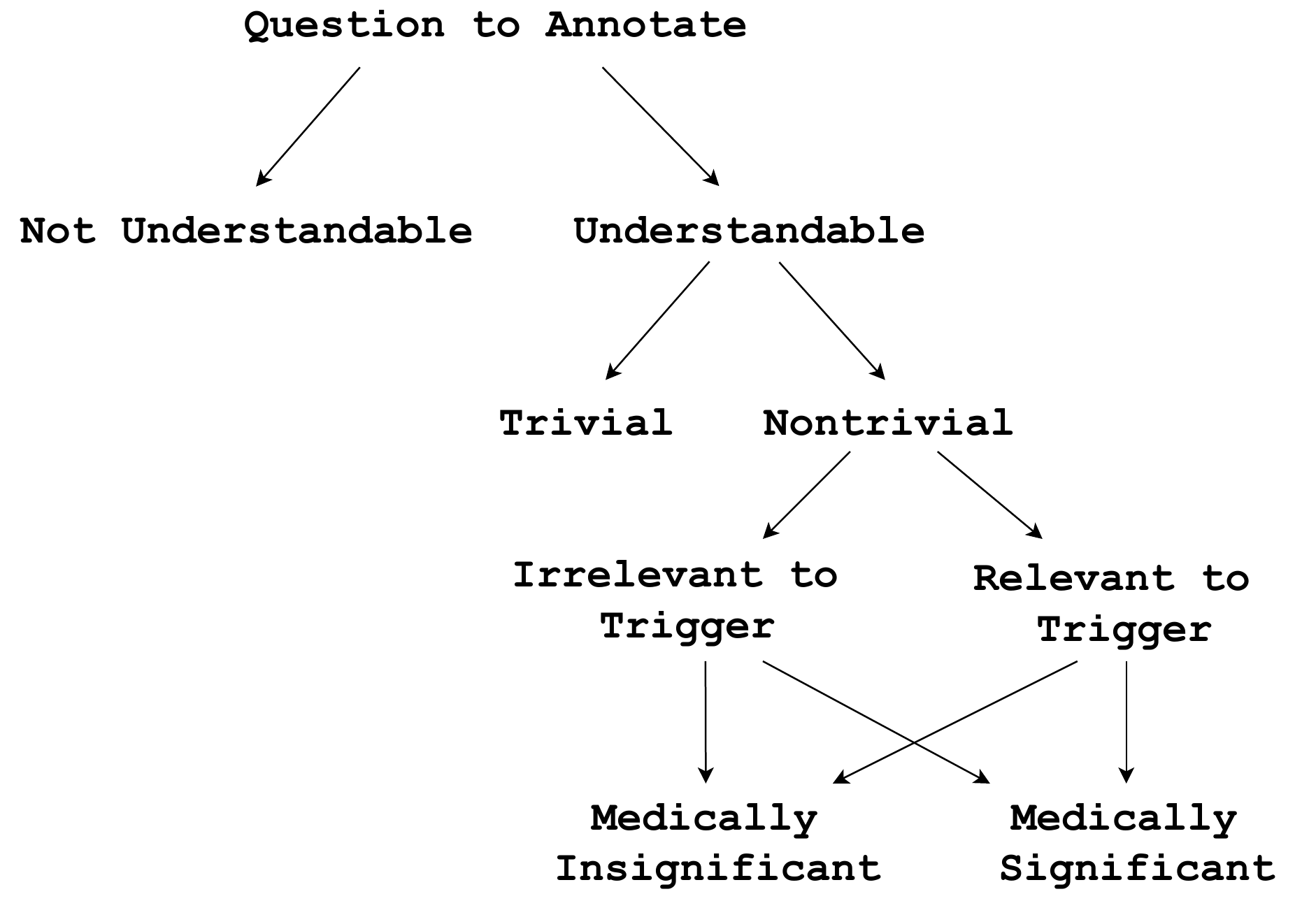}
    \caption{A breakdown of how questions are annotated.}
    \label{fig:annotation_breakdown}
\end{figure}
Human evaluation is still the most reliable way to compare generative models for diverse tasks like question generation.
Common categories for question generation to consider are grammar, difficulty, answerability and fluency \cite{Nema2019LetsAA, tuan2019capturing, wang-etal-2020-answer, Huang_Fu_Mo_Cai_Xu_Li_Li_Leung_2021}. However, not all of these categories are relevant to \textit{clinical} question generation.
We evaluate questions generated using our pipeline, as well as gold standard questions on the following four categories~(binary scale):

\paragraph{Understandability} Can an individual familiar with medical/clinical language understand the information needs expressed, even if the question is not a complete sentence or contains grammar/spelling errors?

\paragraph{Nontriviality} Is the question unanswerable with respect to the sentence it was triggered/generated from? A question that would be considered \textit{trivial} would be ``Did the patient have a fever?'' if the context
presented was ``The patient had a fever''.

\paragraph{Relevancy to trigger} Is the trigger or the sentence containing the trigger related to the question?

\paragraph{Clinical meaningfulness} Will the answer to this question be helpful for further treatment of this patient or understanding the patient’s current condition? Or alternatively, is it reasonable that a medical professional would ask this question given the provided context?

Annotations were divided evenly between medical experts. Each question is scored independently by two different annotators. However, due to time constraints, there are no discussions between annotators about their decisions. We also ensure that annotators did not receive discharge summaries that they had seen previously. Lastly, it is important to note that annotations were assigned blindly. Annotators were informed that they would be scoring both human and machine generated questions, but were not informed about (1) where the question was generated from (i.e., human or machine) and (2) the proportion of human:machine generated questions.

We score questions using the tree presented in Figure \ref{fig:annotation_breakdown}. In cases in which the question is both understandable and nontrivial, we additionally ask medical experts to determine whether or not the proposed answer fully answers, partially answers or is irrelevant to the question. Results can be seen in Table \ref{tab:question_generation_human} and Table \ref{tab:answer_success}.

\begin{table*}[h]
    \centering
    \begin{tabular}{l|rrrrrr}
    \toprule
    Model Type & Context  & Split Qs & Unique Question Ratio & METEOR & BERTScore & ROUGE-L\\
    \midrule
    BART       & Trigger  & N        & 0.301                 & 3.6  & 0.856    & 10.2 \\   
    BART       & Trigger  & Y        & 0.037                 & 0.1  & 0.838    & 3.4 \\ 
    \midrule
    BART       & Sentence & N        & 0.526                 & 6.1  & 0.860    & 10.2 \\   
    BART       & Sentence & Y        & 0.468                 & 7.8  & 0.858    & 12.0 \\   
    BART       & Chunk    & N        & 0.741                 & 7.9  & 0.861    & 11.9 \\   
    BART       & Chunk    & Y        & 0.619                 & 7.2  & 0.861    & 11.6 \\   
    \midrule
    T0-11B & Sentence & N & \textbf{0.779} & 3.9 & 0.861 & 11.9 \\
    T0-11B & Sentence & Y & 0.410 & \textbf{8.4} & \textbf{0.884} & 12.2 \\
    T0-11B & Chunk & N & 0.398 & 3.7 & 0.860 & \textbf{12.4} \\
    T0-11B & Chunk & Y & 0.400 & 6.7 & 0.879 & 10.9 \\
    \end{tabular}
    \caption{Automated metrics for baseline models on the question generation task. \textit{Sentence} and \textit{Chunk} contexts include both the text surrounding the trigger and the trigger itself. \textit{Trigger} context only includes trigger text. Split Qs means splitting multiple questions for a trigger into multiple examples (unique question ratio of these models should not be compared). Results given on dev set.}
    \label{tab:question_generation_automatic}
\end{table*}

\begin{table*}
    \begin{tabular}{l|rrrrr|r}
    \toprule
    Model & Triggers & Understandable & Nontrivial & Relevant &  Clinically Meaningful & Satisfies All \\
    \midrule 
    Gold & -     & 93.8\% & 86.0\% & 83.3\% & 82.3\% & 80.5\% \\
    BART & Gold  & 81.5\% & 59.8\% & 52.3\% & 54.8\% & 47.8\% \\
    T0   & Gold  & 85.8\% & 72.3\% & 68.0\% & 66.5\% & \textbf{62.5\%} \\
    \midrule
    BART & Predicted  & 78.3\% & 57.3\% & 49.3\% & 49.8\% & 41.8\% \\
    T0   & Predicted  & 76.8\% & 49.0\% & 45.0\% & 44.5\% & 41.0\% \\
    \end{tabular}
    \caption{We present results of human evaluation on generated questions. Gold refers to questions generated by medical experts. We do not annotate whether or not a question is  nontrivial, relevant and clinically meaningful if it is not understandable, thus lowering the number of questions that satisfy these categories.}
    \label{tab:question_generation_human}
\end{table*}

\begin{table}[ht]
    \centering
    \begin{tabular}{l|rrrrr|r}
    Model & Triggers & Partially & Fully\\
    \midrule
    Gold & - & 15.0\% & 7.50\% \\
    BART & Gold & 13.75\% & 7.75\% \\
    T0   & Gold & 11.5\% & 6.00\% \\
    BART & Predicted & 14.5\% & 6.25\%\\
    T0   & Predicted & 9.75\% & 3.25\% \\
    \end{tabular}
    \caption{Percent of the time that the answer retrieved by our model partially answers and fully answers the question.}
    \label{tab:answer_success}
\end{table}

\section{Discussion}
We evaluate performance of both the best BART and T0 model with respect to ROUGE-L score. We select 400 questions generated from each model, half of which are generated with gold triggers and the other half with predicted triggers, as described in Section \ref{sec:trigger_dection}. Two medical experts score each question. Due to the subjective nature of the task, we find moderate agreement between annotators with respect to scoring questions ($\kappa = 0.46$) and scoring answer sufficiency ($\kappa = 0.47$). We use the ``Satisfies All'' column (i.e., satisfies all four human evaluation categories) to calculate agreement between questions.

Results show that the T0 model prompted with gold triggers successfully generates a high-quality question 62.5\% of the time (Table \ref{tab:question_generation_human}). This model significantly outperforms BART when given gold-standard triggers. However, the performance significantly drops when the triggers are no longer provided.
We find that T0 produces a large number of \textit{trivial} questions when given a predicted trigger. More testing and investigation is needed to further understand this large drop in performance, as we do not observe this same behavior with BART. 

As human evaluation demonstrates, despite low automatic metric scores, both BART and T0 achieve reasonable success in generating coherent, relevant and clinically interesting questions. To evaluate if the automated metrics can capture the quality of generated questions, we calculate the Spearman’s Rank Correlation Coefficient between human evaluation and automatic metrics. We find extremely low and statistically insignificant correlation for ROUGE-L (-0.09), METEOR (-0.04) and BERTScore (-0.04). This is unsurprising, as these automatic metrics are not designed to capture the categories we examine during human evaluation.

We also score the answers selected by our ClinicalBERT model trained on emrQA (Section \ref{sec:answer_selection}).
Interestingly, we find that of the answers the model successfully recovers, 44\% are extracted from the remainder of the discharge summary used to generate the question. The remaining 56\% come from nursing notes, Radiology/ECG reports and previous discharge summaries. However, for a majority of the questions, we are unable to recover a sufficient answer (Table \ref{tab:answer_success}). We sample 50 gold standard questions whose suggested answers were marked as invalid, in order to determine if this was due to the model's poor performance. We find that 36\% of the questions do in fact have answers in the EHR, thus demonstrating the need for improved clinical QA resources and models.

\section{Conclusion}

We present \textbf{Di}scharge \textbf{S}ummary \textbf{C}linical \textbf{Q}uestions (DiSCQ), a new human-generated clinical question dataset composed of 2000+ questions paired with the snippets of text that prompted each question. 
We train baseline models for trigger detection and question generation.
We find that despite poor performance on automatic metrics, we are able to produce reasonable questions in a majority of cases when given triggers selected by medical experts. 
However, we find that performance significantly drops when given machine predicted triggers.
Further, we find that baseline models trained on emrQA are insufficient for recovering answers to both human and machine generated questions. 
Our results demonstrate that existing machine learning systems, including large-scale neural networks, struggle with the tasks we propose. 
We encourage the community to improve on our baseline models.
We release this dataset and our code to facilitate further research into realistic clinical question answering and generation \href{https://github.com/elehman16/discq}{here}.

\section{Acknowledgements}
This work was supported and sponsored by the MIT-IBM Watson AI Lab. The authors would like to thank Sierra Tseng for feedback on a draft of this manuscript, as well as Melina Young and Maggie Liu for their help in designing some of the figures.

\bibliography{main}
\bibliographystyle{acl_natbib}
\appendix

\section{Appendix}
\label{sec:appendix}
\subsection{Model and Metric Implementation}
To run BART and T0, we make use of the Huggingface implementations \cite{Wolf2019HuggingFacesTS}. We additionally calculate automated metrics for question generation using Huggingface. For calculating Cohen Kappa, precision, recall, and F1 score, we use sklearn \cite{scikit-learn}.

\subsection{Model Hyperparameters}
We use a majority of the default settings provided by the Huggingface library \cite{Wolf2019HuggingFacesTS}. However, we do experiment with varying learning rates (2e-5, 2e-4, 3e-4, 4e-4), warm up steps (100, 200), and weight-decay (0, 1e-6, 1e-3, 1e-1). For the best BART model, we find that using a learning rate of 2e-4, warm up steps of 200, and weight decay of 1e-6 led to the best model. For the T0 model, we find that using a learning rate of 3e-4, running for 100 warmup steps and using a weight-decay of 0.1 led to the best performance. We run for 50 epochs on the BART model and 30 epochs on the T0 model. We use the best epoch with respect to evaluation loss. In our dev set evaluation, we use a beam search width of 5. We use a gradient accumulation step of 32 and 16 for our BART model and T0 model, respectively, 

\subsection{GPUs and Run Time}
For the BART models, we run on 4 GeForce GTX TITAN X. Due to the limited size of these GPUs, we only use a batch size of 1 per GPU. The BART style models take roughly 8 hours to finish training.

For the T0 models, we train using eight V100 GPUs. We set batch size to be 2 per GPU. These models take roughly 24 hours to train. 

\subsection{Risk of Patient Privacy}
We will release our code and data under MIMIC-III access. \citet{Carlini2021ExtractingTD} warns against training large-scale transformer models (particularly ones for generation) on sensitive data. Although MIMIC-III notes consist of deidentified data, we will not release our model weights to the general public. With respect to the trigger detection system, there is less risk in releasing the model weights, as BERT has not been pretrained with generation tasks \cite{Lehman2021DoesBP}. We caution all follow up work to take these privacy concerns into account. 

\end{document}